\documentclass[conference]{IEEEtran}
\usepackage{times}

\usepackage[numbers]{natbib}
\usepackage{multicol}
\usepackage[bookmarks=true]{hyperref}

\usepackage{amsthm}
\usepackage{amsmath,amssymb}
\let\labelindent\relax
\usepackage{enumitem} 
\usepackage{graphics} 
\usepackage{float,url}
\usepackage{graphicx}
\usepackage{subcaption}  

\usepackage{ wasysym } 

\usepackage[ruled,vlined,linesnumbered]{algorithm2e}

\usepackage[usenames, dvipsnames]{color} 

\usepackage{xspace}
\makeatletter
\DeclareRobustCommand\onedot{\futurelet\@let@token\@onedot}
\def\@onedot{\ifx\@let@token.\else.\null\fi\xspace}

\makeatother

\newcommand{\R}{\mathbb{R}} 
\newcommand{\U}{\mathbb{U}} 
\newcommand{\C}{\mathbb{C}} 
\newcommand{\W}{\mathbb{W}} 
\newcommand{\M}{\mathbb{M}} 

\let\oldnl\nl
\newcommand{\nonl}{\renewcommand{\nl}{\let\nl\oldnl}}

\begin{document}

\title{\vspace{-0cm}
\includegraphics[width=\textwidth]{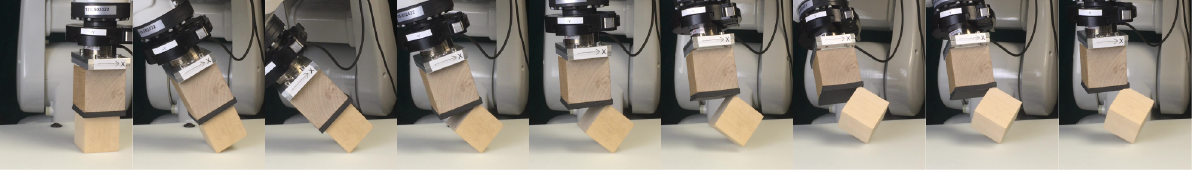}
\centering
\\
 Manipulation with Shared Grasping
}

\author{\authorblockN{Yifan Hou}
\authorblockA{Carnegie Mellon University\\
yifanh@cmu.edu}
\and
\authorblockN{Zhenzhong Jia}
\authorblockA{Southern University of Science and Technology\\
jiazz@sustech.edu.cn}
\and
\authorblockN{Matthew T. Mason}
\authorblockA{Carnegie Mellon University\\
mattmason@cmu.edu}
}


%

\maketitle

\begin{abstract}
A \textit{shared grasp} is a grasp formed by contacts between the manipulated object and both the robot hand and the environment. By trading off hand contacts for environmental contacts, a shared grasp requires fewer contacts with the hand, and enables manipulation even when a full grasp is not possible.
Previous research has used shared grasps for non-prehensile manipulation such as pivoting and tumbling. This paper treats the problem more generally, with methods to select the best shared grasp and robot actions for a desired object motion.
The central issue is to evaluate the feasible contact modes: for each contact, whether that contact will remain active, and whether slip will occur.  Robustness is important. When a contact mode fails, e.g., when a contact is lost, or when unintentional slip occurs, the operation will fail, and in some cases damage may occur.
In this work, we enumerate all feasible contact modes, calculate corresponding controls, and select the most robust candidate.
We can also optimize the contact geometry for robustness.
This paper employs quasi-static analysis of planar rigid bodies with Coulomb friction to derive the algorithms and controls. Finally, we demonstrate the robustness of shared grasping and the use of our methods in representative experiments and examples. The video can be found at https://youtu.be/tyNhJvRYZNk
\end{abstract}

\IEEEpeerreviewmaketitle


\section{Introduction}
\label{sec:intro}

Grasping is widely adopted in automation industry because of its robustness, especially when compared with non-prehensile manipulative actions. A force-closure grasp can neutralize disturbance forces in any direction up to a magnitude determined by the gripping force.
When grasping locations are not available on the object, the robot can only move the object with the help of external force resources \cite{dafle2014extrinsic}, e.g. pivot or tumble an object on a table \cite{aiyama1993pivoting,yoshida2010pivoting,sawasaki1989tumbling}. Unlike grasping, these are usually non-prehensile manipulations which are less robust against disturbance forces.

\citet{salisbury1987whole} introduced the idea of ``whole arm manipulation'', suggesting using all available surfaces on the robot for more ways to manipulate an object. Following this thought, if we treat the environment as another finger of the robot, manipulation under external contacts is like grasping from the object's point of view. We call it \textit{shared grasping}.
With shared grasping, the object can stay in a force closure even when no force closure grasps are available. Under suitable contact modes, the hand can move the object while maintaining the force closure, which opens possibilities for robust manipulation beyond grasping alone.
This view motivates us to design a more general framework for manipulation under external contacts that can:
\begin{itemize}
    \item Analyze the feasibility and robustness of a shared grasping system.
    \item Find robust robot actions to achieve a desired object motion and/or contact mode.
\end{itemize}


Although grasping and shared grasping are conceptually similar, traditional analyses for grasping stability do not work for shared grasping. In grasping, all contact points are assumed to be sticking \cite{murray1994mls}, while in shared grasping all contact modes are considered to allow for more motion possibilities. With multiple possible contact modes, traditional force-based grasping stability analysis \cite{suarez2006grasp} cannot distinguish some contact modes from each other.

To overcome the mode ambiguity, and also to control the exact motion of the shared-grasped object, our approach models the robot action by \textit{Hybrid Force-Velocity Control (HFVC)}. Force control and velocity control are special cases of HFVC. We use the velocity portion of HFVC to filter out undesired contact modes and maintain the desired object velocity.
Note that the traditional force-based grasping analysis cannot model HFVC properly, because the forces in the velocity-controlled directions are not controllable.

In this work, we provide a method to analyze the quasi-static stability of planar shared grasping systems in any contact mode under HFVC. The foundation of our stability analysis is a distinguishable cone representation for the object wrenches from different contact modes. In the object wrench space, we illustrate how the change of robot action and model parameters can cause mode transitions, so as to quantify the \textit{stability margin}, i.e. how much disturbance a contact mode can handle. To get around the force indeterminacy in velocity-controlled directions, we project the wrench space onto the force-controlled subspace and pick a force control there. This is why we call our approach \textit{wrench stamping}. During the computation of stability margin, wrench stamping also computes a robust HFVC action to achieve the desired object motion.

We demonstrate how to use wrench stamping in two problems: 1) How to select a contact mode for a multi-contact manipulation task; 2) Given a contact mode, how to choose finger placement locations on the object. In the first problem, we pick the contact mode with highest stability margin. In the second problem, we compute finger locations by maximizing the stability margin for the given mode. This is possible because we can compute the analytical gradient of the stability margin about the contact geometry parameters. Note that we can also optimize over the environmental contact locations and friction coefficients, which could be useful for system design in industrial applications.

The paper is organized as follows. After introducing related work in Section~\ref{sec:related_work}, we first analyze the feasibility of contact modes in a shared grasping system in Section~\ref{sec:feasibility}. Based on the analysis, we derive the stability margin for any contact mode in Section~\ref{sec:evaluate_robustness}. Section~\ref{sec:algorithms} uses our wrench stamping to solve mode selection and contact point optimization. We provide experimental results in several representative shared grasping problems in Section~\ref{sec:experiments}, and conclude with a discussion on limitations and future work in Section~\ref{sec:conclusion}.


\section{Related Work}
\label{sec:related_work}

\subsection{Manipulation Modeling with External Contacts} 
\label{sub:control_of_manipulation_with_external_contacts}
There is lot of prior work on manipulation with external contacts. By pressing a grasped object against a support, the robot can do in-hand manipulation \cite{terasaki1998motion,holladay2015pivot,Hou2018Fast,prehensile} or transport the object without bearing its full weight \cite{aiyama1993pivoting,yoshida2010pivoting}. Without a grasp, a finger can still move an object by planar pushing \cite{lynch-mason-pushing,zhou2016convex, hogan2016feedback}, pivoting or tumbling \cite{sawasaki1989tumbling,HouICRA19Hybrid}. \citet{dafle2014extrinsic} demonstrated that a simple robot hand can do many tasks by sequencing multiple open loop actions with external contacts.

Several of these examples were prehensile, such as moving a block on a surface by rotating the block about one of its corners \cite{aiyama1993pivoting,yoshida2010pivoting} or edges on the surface \cite{sawasaki1989tumbling}. The object is in a force closure formed by two point-finger contacts and a point or edge table contact. These works computed the conditions of maintaining static balance throughout the motion, including minimal friction coefficient (all contacts were sticking) and range of finger locations. Planar pushing can also be prehensile, examples are rotating a paper card on a table by pressing on it with two fingers \cite{kao1992quasistatic}, and in-hand manipulation with a parallel gripper by external contacts \cite{prehensile,chavan2018hand}.

\subsection{Robust Control of Contact-Rich Manipulation} 
\label{sub:robust_control_of_contact_rich_manipulation}
Manipulation control under external contacts is difficult because a control designed for one contact mode can result in another mode.
Previous work on manipulation control were mainly concerned with switching to the right controller for the right mode. It is popular to do mode scheduling (either planned or hand-coded) before computing a feedback controller \cite{hogan2016feedback,han2019local}.

Using hybrid force-velocity control (HFVC, also called hybrid 'force-position' or 'position-force' control) brings robustness to manipulation, because the force control and velocity control can handle model uncertainties and force disturbances, respectively. There is a long thread of work on HFVC for manipulation in a rigid environment \cite{mason1981compliance,yoshikawa1987dynamic,raibert1981hybrid, west1985method}, and stability analysis for the whole robot-contact system under HFVC \cite{mcclamroch1988feedback}.
\citet{uchiyama1988symmetric} computes HFVC for manipulating an object with two fingertips.

\citet{HouICRA19Hybrid} considered the case when the object may have multiple contacts with both the robot hand and the environment. They introduced the hybrid servoing algorithm to compute a HFVC action, and provided criteria to evaluate the robustness of a multi-contact system against several kinds of failures \cite{Hou2019Criteria}. This work extends previous work by considering all possible contact modes, as well as proposing a simpler robustness score with clear physical meanings.

\subsection{Motion Planning with Contacts} 
\label{sub:motion_planning_with_contacts}
The model of rigid body contacts is discontinuous. Gradients do not exist at the making or breaking of contacts, 
thereby making powerful continuous solvers unusable.
A popular way around is to model the contacts by complementary constraints on continuous variables \cite{posa2014direct}. However, these constraints make the optimization problem close to ill-condition and sensitive to the choice of initial trajectory. Another direction is to approximate the contact dynamics with continuous models \cite{mordatch2012contact,mordatch2012discovery,toussaint2018differentiable}. So far there have been limited success \cite{mordatch2015ensemble} in transferring motions planned with simplified dynamics into experiments.
Sampling based planning methods handle discrete states naturally. They find solutions by solving steering problems under contact constraints \cite{webb2013kinodynamic,chavan2017sampling}, or by projecting the sampled motion into the constrained manifold \cite{berenson2009manipulation}.

A key to successful experiments is the ability to handle uncertainties. Object pose uncertainties can be reduced by contacts \cite{Koval-2016-5594,zhou2017probabilistic,sieverling2017interleaving}. However, there is little work on making the contact modes robust against modeling uncertainties and force disturbances. This paper lays the foundation for such planning by providing fast solutions for robustness-based mode selection and contact point selection.

In recent years, learning based methods have also enabled robots to do manipulation tasks with external contacts, such as opening doors \cite{gu2017deep} and pushing in clutter for grasping \cite{finn2017deep,zeng2018learning}. However, these learned tasks are intrinsically stable: temporarily falling into a bad contact mode will not fail the task. We have not seen work on learning shared grasping tasks such as pivoting or tumbling. The reason could be the difficulty in simulating contacts and the inefficiency/danger of action exploration under holonomic constraints.

\section{Feasibility of a Contact Mode}
\label{sec:feasibility}
To begin with, in this section we derive the mechanical condition under which a contact mode is feasible and unique. This section lies the base for our robustness analysis in the next section. First, we introduce the symbols.

Consider a robot hand, a rigid object and a rigid environment. Denote $H, O, W$ as the hand frame, object frame and world inertia frame, respectively. We use the symbol $^Ac_B$ to describe the quantity $c$ of frame $B$ measured in frame $A$, where $c$ could be position $p$, normal $n$, force $f$, etc.
Denote $N_D$ as the degree-of-freedom of a rigid body, $N_d$ as the dimension of space ($N_d=2$ for planar systems). We make the following assumptions:
\begin{itemize}
  \item The system is quasi-static: inertia forces are ignored.
  \item Friction obeys Coulomb's Law; the maximum stiction equals the sliding friction.
  \item The hand joints are locked.
  \item Gravity is not significant comparing to contact forces.
\end{itemize}
A description of a shared grasping system includes:
\begin{itemize}
  \item Robot hand pose and object pose.
  \item A list of contact points and normals on the object.
  \item Coulomb friction coefficient $\mu$ for each contact.
\end{itemize}

\subsection{Newton's Law for Wrench Space Analysis} 
\label{sub:Newtons_law}
Define the generalized velocity vector $V$ and force vector $F$ of the system in the robot hand frame $H$:
\begin{equation}
V = \left[\begin{aligned}^Hv_O\\^Hv_H\end{aligned}\right],
F = \left[\begin{aligned}^Hf_O\\^Hf_H\end{aligned}\right].
\end{equation}
$^Hf_H$ is the robot control force, $^Hf_O$ is always zero since the object is free.
We choose the hand frame because it is easier to describe HFVC later. The Newton's Second Law for the hand-object system takes the form:
\begin{equation}
  \label{eq:newtons law}
  \left[ {\begin{array}{*{20}{c}}
{{\bf J}^T_e}&{-{\bf J}^T_h}\\{0}&{{\bf J}^T_h}\end{array}} \right]
\left[\begin{aligned}\tau_e\\\tau_h\end{aligned}\right] +
\left[\begin{aligned}^Hf_O\\^Hf_H\end{aligned}\right]
=0.
\end{equation}
where $\tau_e, \tau_h \ge 0$ are vectors of contact wrench magnitudes; each row of the contact Jacobian ${\bf J}_e$ and ${\bf J}_h$ represents the wrench of a friction cone edge of an environment contact and hand contact, respectively. The derivation of the contact Jacobians can be found in \cite{murray1994mls}. Each contact point contributes up to two rows to the contact Jacobian, depending on its mode:
\begin{itemize}[] 
  \item[] (\textbf{s}) Separation: no wrench from the contact;
  \item[] (\textbf{f}) Fixed: wrenches for both the left and right edges;
  \item[] (\textbf{l}) Left sliding: only the wrench for the right edge;
  \item[] (\textbf{r}) Right sliding: only the wrench for the left edge.
\end{itemize}
Following the convention in \cite{mason2001mechanics}, we use a string of the abbreviation letters to describe a contact mode, e.g. `ffsl'.

\subsection{A Distinguishable Cone Representation for Modes} 
\label{sub:cone_intersection}
There are many ways to check the feasibility of ~\eqref{eq:newtons law}, i.e. whether a set of feasible forces exist. It is well-known that all the contact wrenches on the object forms a polyhedral convex cone (PCC) in the object wrench space; if the cone spans the whole space, the contacts on the object are in static balance and the object is in force closure \cite{murray1994mls}. This way of using PCC is not helpful for shared grasping, because it cannot distinguish different contact modes and every force closure contact mode looks the same (span the whole wrench space). We must modify the representation.

\begin{figure}[ht]
    \centering
    \includegraphics[width=0.45\textwidth]{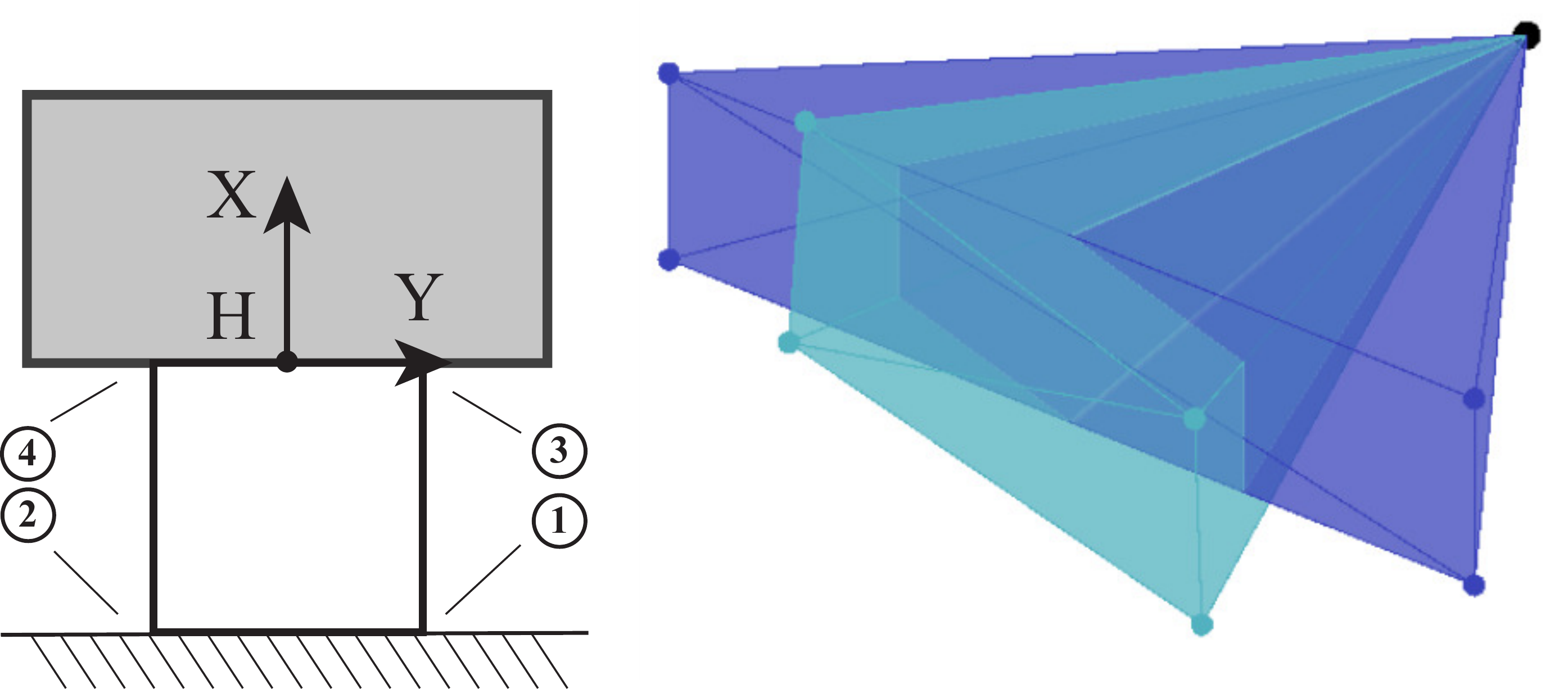}
    \caption{Left: a shared grasping system with a cube object (bottom) and a robot palm (top). The circled numbers show the ordering of contact points. Right: the hand cone (purple) and environment cone (blue) for the ``ffff'' mode.}
    \label{fig:example1}
\end{figure}

Rewrite equation~\eqref{eq:newtons law} using $^Hf_O=0$:
\begin{equation}
  \label{eq: cone Jh Je f}
  {\bf J}^T_e\tau_e = {\bf J}^T_h\tau_h = -{^Hf}_H.
\end{equation}
This equation maps the environmental contact force $\tau_e$ and the hand contact force $\tau_h$ into the generalized force space of the hand, which is the space where the robot action ${^Hf}_H$ lives in.
Since $\tau_e \ge 0, \tau_h \ge 0$, ${\bf J}^T_e\tau_e$ and ${\bf J}^T_h\tau_h$ each again forms a polyhedral convex cone in this wrench space, call them the \textit{environment cone} $\C_e$ and the \textit{hand cone} $\C_h$:
\begin{equation}
\label{eq:h cone and e cone}
  \begin{aligned}
  \C_e = \{{\bf J}^T_e\tau_e\ |\ \tau_e\ge0\} \\
  \C_h = \{{\bf J}^T_h\tau_h\ |\ \tau_h\ge0\}
  \end{aligned}
\end{equation}
An example is shown in Figure~\ref{fig:example1}.
Rows of ${\bf J}_e$ and ${\bf J}_h$ are generators of the cones. Our condition of force balance is:

\textit{The existence of a nonzero intersection between $\C_e$ and $\C_h$ is equivalent to the existence of a non-trivial (all zero) solution $(\tau_e, \tau_h, {^Hf_H})$ to the static balance equation \eqref{eq: cone Jh Je f}.}

\begin{figure}[ht]
    \centering
    \includegraphics[width=0.35\textwidth]{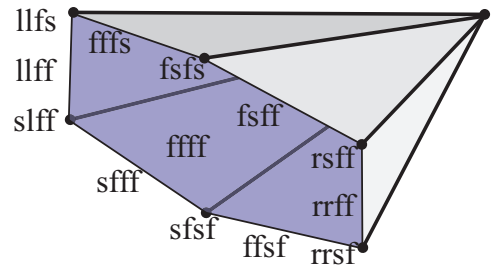}
    \caption{A closer look at the cones of the example in Figure~\ref{fig:example1} with annotation. In this example, each non-zero face of the 3D cone is the cone of a contact mode.}
    \label{fig:all_cones}
\end{figure}

Now we explain how this new intersection-based condition makes a difference in distinguishing contact modes. We denote this intersection as $\C_m$ and call it the \textit{cone of the mode $m$}. We can draw $\C_m$ for all the modes in the same space, as in Figure~\ref{fig:all_cones}. We use the contact mode enumeration algorithm by \citet{huang2020modes} to get a list of modes.
Note that these cones are distinguishable, although most contact modes have force closure. This is because different contact modes always have different contact wrenches.

From \eqref{eq: cone Jh Je f} we know the robot can directly apply a wrench $w=-{^Hf}_H$ in this wrench space, then as long as
\begin{equation}
  w\in\C_m,
\end{equation}
mode $m$ is feasible under robot action $w$. We can make the claim stronger:

\textit{If the robot action $w$ is contained only in $\C_m$, the system must be in mode $m$.}

Now it seems straightforward to enforce a contact mode $m$: pick a robot wrench $w$ in $\C_m$ while staying away from the cones of other modes. The system will be in mode $m$ if such $w$ exists. However, this approach lacks robustness because the cones of different modes are not well separated from each other. For example, in Figure~\ref{fig:all_cones} every facet of the ``ffff'' mode denotes another mode. If we apply $w$ in one of those modes, a tiny disturbance force could push the actual applied wrench into the ``ffff'' mode. This shows a limitation of purely force-based method in multi-contact manipulation. We show how to overcome it in section~\ref{sub:hybrid_force_velocity_control} shortly.

Before we move on, another thing we can learn is the range of wrenches the robot may apply. Denote $\U_{\rm AF}$ as the union of the hand cone and environment cone of the all-fixed mode. Denote $\W$ as the whole $N_D$ dimensional wrench space. We can tell the result of applying a wrench $w$ as follows.
\begin{itemize}
  \item $w\in \C_{\rm AF}$: the object can have static force balance;
  \item $w\in \U_{\rm AF} - \C_{AF}$: the contact forces cannot balance the wrench, the object is accelerating;
  \item $w\in \W - \U_{\rm AF}$: (impossible) some contact forces are outside of their friction cones.
\end{itemize}
Depending on the existence of $\C_m$, we call mode $m$ \textit{F-feasible} or \textit{F-infeasible}.

\subsection{Feasibility under Hybrid Force-Velocity Control(HFVC)} 
\label{sub:hybrid_force_velocity_control}
In this subsection we explain how to uniquely select one mode by replacing pure force control $w$ with HFVC. HFVC decompose the robot action space into a velocity control subspace and force control subspace. Denote $n_{af}$ and $n_{av}$ as the dimension of force and velocity control subspace in HFVC, $n_{af} + n_{av} = N_d$. Define transformation ${\bf R}_a\in\R^{N_d\times N_d}$, we can describe a HFVC by:
\begin{equation}
\label{eq:hfvc}
  \begin{aligned}
    \left[ {\begin{array}{*{20}{c}}{{\omega _{af}}}\\{{\omega _{av}}}\end{array}} \right] = {{\bf R}_a}{^Hv}_H,\quad
    \left[ {\begin{array}{*{20}{c}}{{\eta _{af}}}\\{{\eta _{av}}}\end{array}} \right] = {{\bf R}_a}{^Hf}_H,
  \end{aligned}
\end{equation}
where $\eta_{af}\in\R^{n_{af}}$ and $\omega_{av}\in\R^{n_{av}}$ are the force and velocity control, respectively. $\eta_{av}$, $\omega_{af}$ denote the uncontrollable variables: force/velocity in the velocity/force-controlled direction.
The HFVC imposes a velocity constraint:
\begin{equation}
\label{eq:v command}
  \left[ {\begin{array}{*{20}{c}}
0&{{\bf R}_a^{({n_{av}})}}
\end{array}} \right]V = {\omega_{av}}.
\end{equation}
${\bf R}_a^{({n_{av}})}$ is the last $n_{av}$ rows of ${\bf R}_a$. Denote ${\bf C}_v=[0\ \ {\bf R}_a^{(n_{av})}]$ and write the velocity constraint as ${\bf C}_vV=\omega_{av}$.

Given a desired contact mode, we use hybrid servoing \cite{HouICRA19Hybrid} to compute the force-velocity decomposition ${\bf R}_a$ and velocity control $\omega_{av}$. Hybrid servoing quickly finds a velocity control that is compatible with the contact constraints. We don't use the second half of hybrid servoing which computes the force control $\eta_{af}$, because the original algorithm was not designed for handling sliding contacts. The inputs to hybrid servoing include a description of the goal velocity:
\begin{equation}
  \label{eq:velocity goal}
  {\bf G}V=b_G,
\end{equation}
and contact constraints on velocity ${\bf N}_mV=0$, as explained below. In the following we analyze the system under a HFVC and use a new approach to compute the force control $\eta_{af}$.

\subsubsection{Model filtering by velocity control} 
\label{subsub:model_filtering_by_velocity_control}
Denote the velocity constraint imposed by the contacts in contact mode $m$ by:
\begin{equation}
  \label{eq:contact constraint}
  \begin{array}{*{20}{c}}
      {\bf N}_mV= 0 \\
      {\bf M}_mV\le 0,
  \end{array}
\end{equation}
including non-penetration constraint, sticking constraint and sliding direction constraint. We call mode $m$ \textbf{V-feasible} if the velocity constraints \eqref{eq:v command} and \eqref{eq:contact constraint} have a solution, \textbf{V-infeasible} otherwise.

Before we can safely ignore all the V-infeasible modes, we need to consider the possibility of \textit{crashing} \cite{Hou2019Criteria}, i.e. the situation where a small robot motion causes huge internal force. Crashing can be dangerous and must be avoided. In crashing, the system can remain in static balance when the force in the velocity-controlled directions becomes arbitrarily large. In other words, $\C_m$ must intersect the set of positive forces in the velocity-controlled directions, denoted as $\C_v$:
\begin{equation}
  \label{eq:crashing condition}
  \C_m \cap \C_v \ne \{0\}.
\end{equation}
This is a necessary condition. $\C_v$ here can be derived from \eqref{eq:h cone and e cone} and \eqref{eq:hfvc}:
\begin{equation}
  \label{eq:velocity cone}
  \C_v = \{-{\bf R}^{-1}_{av}\tau_v | \tau_v\ge0\},
\end{equation}
where ${\bf R}^{-1}_{av}$ represents the last $n_{av}$ columns of ${\bf R}^{-1}_{a}$. $\C_v$ is again a PCC, we call it the \textit{velocity cone}.

Remember $\C_{AF}$ contains all $\C_m$. Using \eqref{eq:crashing condition} we can derive the sufficient condition for crashing-avoidance:
\begin{equation}
  \label{eq:crashing condition AF}
  \C_{AF} \cap \C_v = \{0\},
\end{equation}
This condition ensures no crashing in all possible modes. If it is satisfied, we can safely ignore all the V-infeasible modes, which makes it easier to separate the desired mode from the rest. This is illustrated in the middle figure of Figure~\ref{fig:wrench stamping}.

\subsection{Mode Selection by Force Control} 
\label{sub:mode_selection_by_force_control}
There is usually more than one V-feasible mode. To secure the desired mode, the robot still needs to apply suitable force.
\begin{figure}[ht]
\centering
\includegraphics[width=0.45\textwidth]{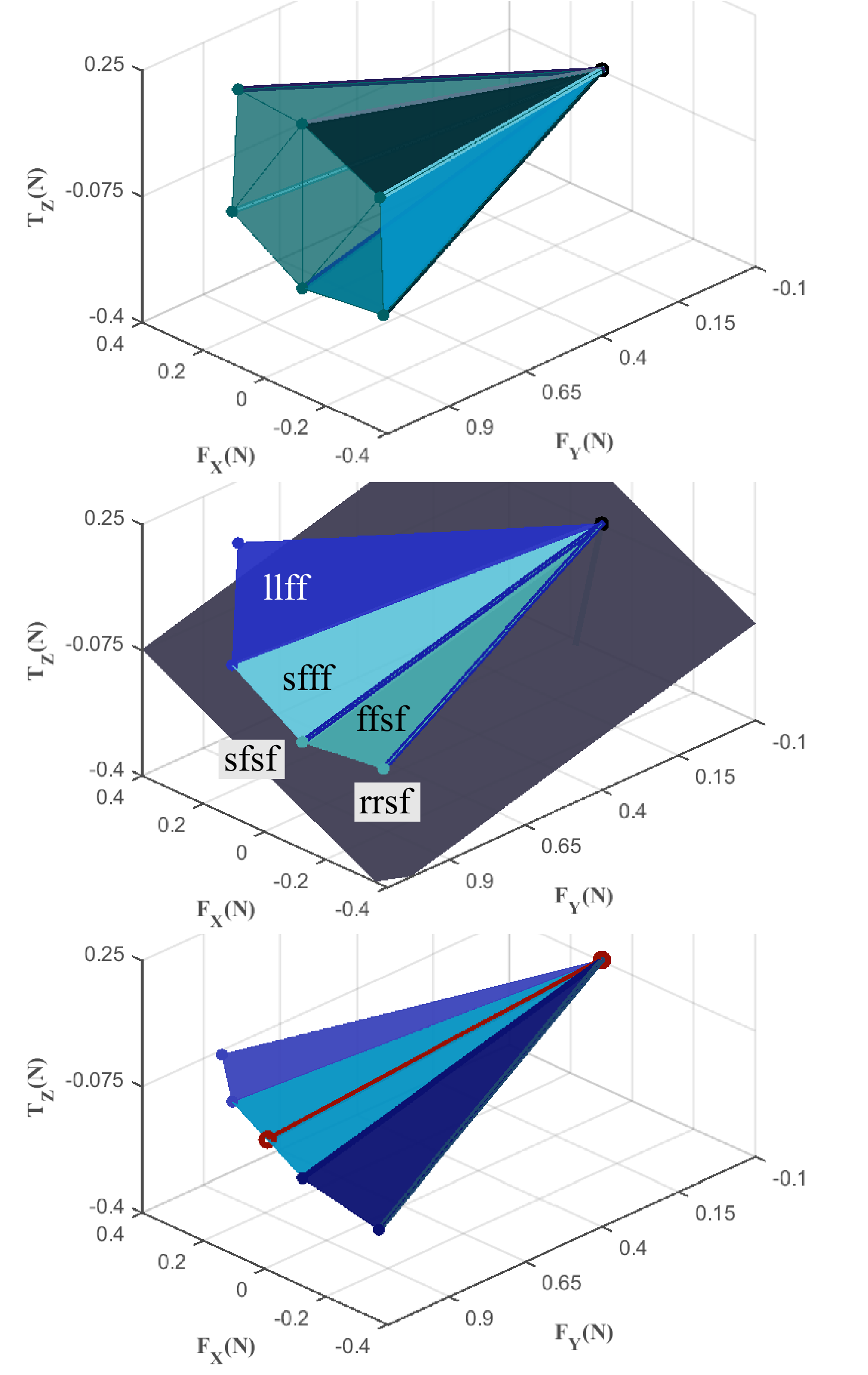}
\caption{Procedures of wrench stamping for the problem in Figure~\ref{fig:example1}. The torque is scaled into Newton by the object length. Top: All nonempty cones. Middle: the cones of V-feasible modes (one or two generators). The gray plane is the force-controlled subspace. Bottom: Projection of feasible cones onto the force-controlled subspace. The red ray is the chosen wrench for mode ``sfff''.}
\label{fig:wrench stamping}
\end{figure}
Equation~\eqref{eq:hfvc} and \eqref{eq: cone Jh Je f} gives:
\begin{equation}
  \label{eq: cone Jh Je eta}
  {\bf J}^T_e\tau_e  = {\bf J}^T_h\tau_h = w = - {\bf R}^{-1}_a\left[\begin{aligned}\eta_{af}\\\eta_{av}\end{aligned}\right],
\end{equation}
so the force-velocity decomposition happens directly in the wrench space where the hand and environmental cones live.

Note again that we can \textit{not} control $\eta_{av}$, so we cannot directly place the wrench vector $w$ in the cone of the desired mode. To resolve the difficulty, we project all V-feasible cones $\C_m$ into the force-controlled subspace. Those projections $\bar \C_m$ are PCCs in a lower dimensional space. Then we pick the force control $\eta_{af}$ to be within the projection of the desired mode, while staying away from any other projections:
\begin{equation}
  \label{eq:choice of force}
  \eta_{af} \in \bar\C_{\rm goal} - \bar\C_{\rm others}.
\end{equation}
The minus sign means set difference. If $\eta_{af}$ exists, the actual $w$ has to be in the desired mode because projections are surjective. An example of is shown in Figure~\ref{fig:wrench stamping}, bottom. Because of the projection, we call this approach \textit{Wrench Stamping}.

If the set $\bar\C_{\rm goal} - \bar\C_{\rm others}$ is empty, we call this mode \textit{F-indistinguishable} since there is no force action the robot can take to avoid other feasible modes.

\section{Robustness of a Contact Mode} 
\label{sec:evaluate_robustness}
In this section, we show how to evaluate the robustness of a contact mode by looking at two causes of mode transitions:
\subsection{Perturbations on Contact Geometry} 
\label{sub:geometry}
Contact geometries plus friction coefficients determine the shape of hand and environment cones $\C_e, \C_h$. Modeling uncertainties change the shape of these cones, which could cause their intersection to disappear and the corresponding mode to be F-infeasible.  For planar systems, we define a stability margin $\Phi_g(\C_e, \C_h)$ that describes the ``depth'' of the intersection, i.e. the minimal angular rotation $\C_e$ needs to take to eliminate non-trivial intersections with $\C_h$. Note that the volume (solid angle) of the intersection $\C_m$ is NOT the right metric. Cones of most modes have zero volume. To introduce our metric, we first define a distance function $\Delta(F, E)$ which computes the angular distance between ray $E$ and hyperplane $F$:
\begin{equation}
  \label{eq:distance function}
  \Delta(F, E) = \arcsin(n_F\cdot E),
\end{equation}
where $n_F$ denotes the normal of the hyperplane $F$. Both $n_F$ and $E$ are unit vectors. Next, define a function that evaluates the ``depth'' of a cone inside another cone. Consider two PCCs $\C_a, \C_b$ that satisfies $\C_b\subseteq\C_a$. Define
\begin{equation}
  \label{eq: cone depth}
  \sigma(\C_a, \C_b) = \min\limits_{j} \max\limits_{i} \Delta(F_j, E_i),
\end{equation}
where $F_j$ is the $j$th facet of $\C_a$, $E_i$ is the $i$th edge of $\C_b$.
A facet $F_j$ defines a hyperplane with the positive normal pointing inside of the cone. We reduce dimension whenever possible, for example, if $\C_a$ has two generators, a facet only has one ray and defines a one-dimensional hyperplane. If $\C_a$ only has one generator, we define $\sigma(\C_a, \C_b)=0$. Finally, define the \textit{geometrical stability margin} $\Phi_g(\C_e, \C_h)$ as:
\begin{equation}
  \label{eq: stability margin}
  \Phi_g(\C_e,\C_h) = \min (\sigma(\C_e, \C_e\cap\C_h),\ \sigma(\C_h, \C_e\cap\C_h)).
\end{equation}

\subsection{Perturbations on Force Control}
\label{subsub:force_control}
\begin{figure}[ht]
\centering
\includegraphics[width=0.45\textwidth]{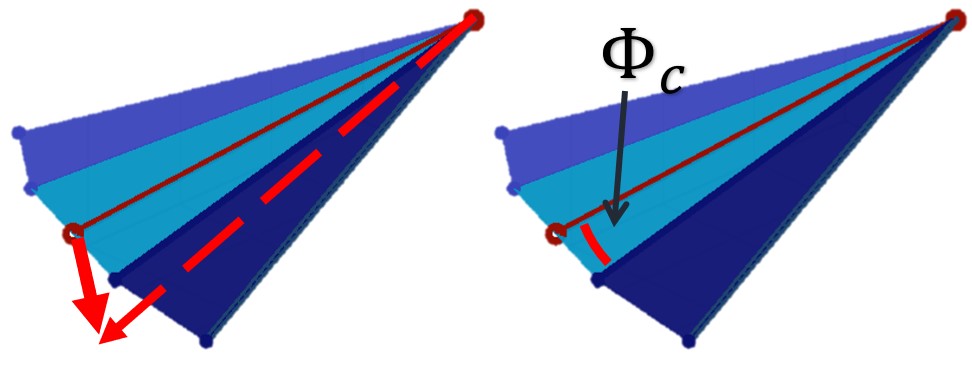}
\caption{Left: A disturbance force (bold red arrow) changes the contact mode. Right: illustration of the control stability margin.}
\label{fig:control stability margin}
\end{figure}

Disturbance forces can move the force control $\eta_{af}$ out of the region defined by equation~\eqref{eq:choice of force}.
We define the \textit{control stability margin} $\Phi_c$ as the angular distance from $\eta_{af}$ to the closest cone projections of other modes $\bar\C_{\rm others}$. Figure~\ref{fig:control stability margin} shows how the control in Figure~\ref{fig:wrench stamping} can fail under disturbance force.

$\Phi_c$ and $\Phi_g$ each describes the minimal disturbance force required to trigger mode transition in terms of angular distance in object wrench space. Define the \textit{stability margin} $\Psi$:
\begin{equation}
  \label{eq:stability margin psi}
  \Psi = \min (\Phi_c,\ \Phi_g).
\end{equation}
With proper scaling between force and torque, as well as a nominal force magnitude $K_F$, the product $K_F\Psi$ represents the minimal magnitude of the disturbance force that can change contact mode $m$. Just like grasping, we can make the system more robust simply by applying larger force.

We summarize wrench stamping in Procedure~\ref{proc:wrench stamping}. Line~\ref{line:mode enumeration} of \ref{proc:wrench stamping} calls 2D contact mode enumeration algorithm implemented in \cite{huang2020modes}. Line~\ref{line:hybrid servoing} uses the first half of the hybrid servoing algorithm \cite{HouICRA19Hybrid}.

\begin{procedure}
\caption{WrenchStamping(Mode $m^*$)}\label{proc:wrench stamping}
\SetAlgoLined
\KwIn{${\bf N}, {\bf M}, \C_e, \C_h$ for all-fixed mode \eqref{eq:contact constraint}\eqref{eq:h cone and e cone}}
\KwIn{$G, b_G$, set of modes $\M$}
\KwOut{$\Psi$, HFVC ($n_{af}, n_{av}, {\bf R}_a, w_{av}, \eta_{af}$)}
$[n_{af}, n_{av}, {\bf R}_a, w_{av}]$ = hybridServoing(${\bf N}_{m^*}, {\bf G}, b_G$)\;\label{line:hybrid servoing}
\lIf{Crashing happens (\eqref{eq:crashing condition AF}\eqref{eq:velocity cone})}{\Return{$\clock$}} \label{line:crashing check}
Initialize empty set ${\hat \M} \leftarrow \clock$\;
\tcp{Mode filtering by F-feasibility}
\ForEach{$m\in \M$}{\label{line:F loop begin}
    Assemble $\C^{(m)}_e$,$\C^{(m)}_h$ from columns of $\C_e,\C_h$\;
    \lIf{$\C^{(m)}_e \cap \C^{(m)}_h = \clock$}{Continue}
    Compute stability margin $\Phi^{(m)}_g(\C^{(m)}_e,\C^{(m)}_h)$ (\ref{eq: stability margin})\;\label{line:compute phi_g}
    \lIf{$\Phi^{(m)}_g=0$}{Continue}
    Assemble ${\bf N}_m,{\bf M}_m$ from rows of ${\bf N}, {\bf M}$\;
    $\hat \M \leftarrow \{\hat \M, m\}$\;
}\label{line:F loop end}
$\M\leftarrow \hat \M, \hat \M \leftarrow \clock$\;
\tcp{Mode filtering by V-feasibility}
\ForEach{$m\in \M, m\ne m^*$}{
    \lIf{\eqref{eq:v command}\eqref{eq:contact constraint} are feasible}{${\hat \M} \leftarrow \{{\hat \M}, m\}$}
}
\tcp{Projections}
$\bar \C_{\rm goal} \leftarrow {\bf R}^{-T}_{af}\C_m$, ${\bf R}^{-1}_{af}$ is the first $n_{af}$ cols of ${\bf R}^{-1}_a$\;
$\bar\C_{\rm others}\leftarrow \clock$\;
\ForEach{$m\in \hat \M$}{
  $\bar \C_{\rm others} \leftarrow \{\bar \C_{\rm others}, {\bf R}^{-T}_{af}\C_m\}$
}
$\bar\C_{\rm remain} \leftarrow \bar \C_{\rm goal} - \bar \C_{\rm others}$\;
\lIf{$\bar\C_{\rm remain}=\clock$}{\Return{$\clock$}}
Pick $\eta_{af}\in \bar\C_{\rm remain}$, compute $\Phi_c$\;
$\Psi \leftarrow \min (\Phi_c,\ \Phi_g)$ ($\Phi_g$ was computed in line~\ref{line:compute phi_g})\;
\end{procedure}


\section{Algorithms for two problems} 
\label{sec:algorithms}
In this section, we provide algorithms for two problems in shared grasping.
\subsection{Shared Grasping Control with Mode Selection} 
\label{sub:shared_grasping_with_mode_selection}
We provide algorithm \ref{alg:shared grasping control} to solve robust control for shared grasping problem. Given a desired velocity, the algorithm finds the most robust contact mode and computes a hybrid force-velocity control.

If a desired contact mode is given, the control problem becomes a hybrid servoing problem \cite{HouICRA19Hybrid}. Procedure~\ref{proc:wrench stamping} itself can solve it better than the original hybrid servoing algorithm: users do not need to provide the guard conditions, \ref{proc:wrench stamping} determines the condition of mode transition by itself. Also \ref{proc:wrench stamping} can handle robust control under sliding contacts.
\begin{algorithm}
\caption{Shared Grasping Control}\label{alg:shared grasping control}
\SetAlgoLined
\KwIn{${\bf N}, {\bf M}, \C_e, \C_h$ for all-fixed mode \eqref{eq:contact constraint}\eqref{eq:h cone and e cone},$G, b_G$}
\KwOut{The chosen contact mode $m$, stability margin $\Psi$, HFVC ($n_{af}, n_{av}, R_a, w_{av}, \eta_{af}$)}
Compute the cone of all-fixed mode: $\C_{\rm AF} \leftarrow \C_e\cap\C_h$\;
$\M_{\rm all}\leftarrow$ all contact modes, $\M \leftarrow \clock$\;\label{line:mode enumeration}
Line~\ref{line:F loop begin} - \ref{line:F loop end} of Procedure~\ref{proc:wrench stamping} for $m\in \M_{\rm all}$, save results in $\M$\;
Initialize the solution set: ${\bf S}\leftarrow\clock$\;
\ForEach{$m\in \M$}{
    ${\bf S}\leftarrow\{{\bf S}, $ \ref{proc:wrench stamping}($m$) $\}$\label{line:call alg 2}
}
\Return{the $s\in {\bf S}$ with highest stability margin $\Psi$}
\end{algorithm}

\subsection{Geometry Optimization} 
\label{sub:geometry_optimization}
We can optimize contact geometry parameters to maximize the geometrical stability margin $\Phi_g$ for a given mode. Note that the function $\Phi_g(\C_e,\C_h)$ boils down to the $\Delta$ distance between a facet $F$ and an edge $E$. The facet has either two generators $C_i$, $C_j$ or just $C_i$, we can backtrack its location in $\C_e$ and $\C_h$. The edge is a generator of $\C_e\cap\C_h$, which could be either an edge $C_k$ of $\C_e$ or $\C_h$, or an intersection of a facet $(C_1, C_2)$ in $\C_e$ and a facet $(C_3, C_4)$ in $\C_h$. In the latter case, the expression of the edge is $(C_1\times C_2)\times(C_3\times C_4)$. So we can explicitly write down $\Phi_g$ as a function of contact screws:
\begin{equation}
  \label{eq: stability margin as contact screw 1D}
  \begin{aligned}
  \Phi_g&(\C_e,\C_h) = \arccos\left( \frac{{{C_i}}}{{||{C_i}||}}\cdot{ \frac{C_k}{||C_k||}} \right) \\
  &{\rm or}\\
  =&\arccos\left( {\frac{{{C_i} }}{{||{C_i} ||}}\cdot\frac{{({C_1} \times {C_2}) \times ({C_3} \times {C_4})}}{{||({C_1} \times {C_2}) \times ({C_3} \times {C_4})||}}} \right).
  \end{aligned}
\end{equation}
for one-dimensional facet $F$, or
\begin{equation}
  \label{eq: stability margin as contact screw}
  \begin{aligned}
  \Phi_g&(\C_e,\C_h) = \arcsin\left( \frac{{{C_i} \times {C_j}}}{{||{C_i} \times {C_j}||}}\cdot{ \frac{C_k}{||C_k||}} \right) \\
  &{\rm or}\\
  =&\arcsin\left( {\frac{{{C_i} \times {C_j}}}{{||{C_i} \times {C_j}||}}\cdot\frac{{({C_1} \times {C_2}) \times ({C_3} \times {C_4})}}{{||({C_1} \times {C_2}) \times ({C_3} \times {C_4})||}}} \right)
  \end{aligned}
\end{equation}
for two-dimensional $F$. We can further expand the contact screws into contact geometry parameters and compute the gradient of $\Phi_g$ in analytical form.
We outline a gradient descent algorithm in Algorithm~\ref{alg:geometrical parameter optimization}. The parameter $\bf p$ may contain contact locations, normals and friction coefficients. Note that computation in (\ref{eq: stability margin as contact screw 1D}) and (\ref{eq: stability margin as contact screw}) only involves contact screws that are the current bottleneck of stability margin. This set of contact screws will change between iterations. We observed this phenomenon in the experiments section.

In order to get analytical expression of the gradient of $\Phi_g$, we ignore the normalizations in (\ref{eq: stability margin as contact screw 1D}) and (\ref{eq: stability margin as contact screw}) when computing their derivatives. To obtain the correct gradient about a contact screw, we just project the gradient computed this way onto the tangent plane of the contact screw, as shown in line~\ref{line:project normalize}.

\begin{algorithm}
\caption{Geometrical Parameter Optimization} \label{alg:geometrical parameter optimization}
\SetAlgoLined
\KwIn{Contact geometry parameter $\bf p$}
\KwOut{Optimized contact geometry parameter $\bf p^*$}
\Repeat{converge}{
    \tcp{1. Evaluate $\Phi_g$}
    Compute $\C_e$, $\C_h$ (\ref{eq:h cone and e cone}) using $\bf p$\;
    Compute $\C_m = \C_e\cap\C_h$, record the mapping $M_1$ between columns of $\C_m$ and columns of $\C_e, \C_h$.\;
    Compute $\Phi_g$ (\ref{eq: stability margin}), record the mapping $M_2$ between $C_{i,j,k,1,2,3,4}$ and columns of $\C_e, \C_h, \C_m$\;
    Use $M_1$ and $M_2$ to compute the mapping $M_3$ between $C_{i,j,k,1,2,3,4}$ and columns of $\C_e, \C_h$\;
    \tcp{2. Compute gradient}
    Compute $\Phi_t = \partial\Phi_g/\partial C_t,  t\in\{i,j,k,1,2,3,4\}$ by differentiating (\ref{eq: stability margin as contact screw 1D})(\ref{eq: stability margin as contact screw}) without denominators\;
    Remove the components of $\Phi_t$ that scales $C_t$:\\
    \nonl \ \ \ \ $\Phi_t = \Phi_t - \Phi_t\cdot C_t, t\in\{i,j,k,1,2,3,4\}$\;\label{line:project normalize}
    Use $M_4$ to construct $\Phi_C = \partial\Phi_g/\partial C$ from $\Phi_t$, $C\in$ columns of $\C_e,\C_h$\;
    Compute the Jacobian $C_{\bf p} = \partial C/\partial {\bf p}$ by differentiating the expression of $\C_e,\C_h$ (\ref{eq:h cone and e cone}).\;
    $\Phi_{\bf p} \leftarrow \sum_{C}(\Phi_C\cdot C_{\bf p})$\;
    \tcp{3. Update}
    ${\bf p} \leftarrow {\bf p} + d \Phi_{\bf p}$, $d$ is a step length\;
}
\end{algorithm}


\section{EXPERIMENTS} 
\label{sec:experiments}
We implement our algorithms in MATLAB and test them on several tasks in experiments. We use an ABB IRB120 robot in section~\ref{sub:robust_control_of_desired_contact_modes} and \ref{sub:contact_mode_selection_and_control}, a UR5e robot in section~\ref{sub:range_of_feasible_geometries}, both are 6-axis robot and position-controlled; we implement a hybrid force-velocity controller with a wrist-mounted ATI Mini-40 FT sensor. The communication rate is 250Hz for the ABB robot and 500Hz for the UR5e robot. In both cases, the force control bandwidth is roughly 1Hz.

\subsection{Robust Control of Desired Contact Modes} 
\label{sub:robust_control_of_desired_contact_modes}
\begin{figure}[ht]
    \centering
    \includegraphics[width=0.45\textwidth]{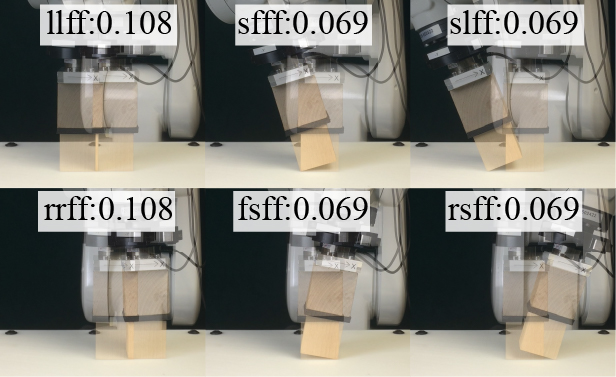}
    \caption{Different contact modes and their stability margins. The order of contacts follows Figure~\ref{fig:example1}.}
    \label{fig:exp_all_modes}
\end{figure}
First, we validate the ability of our method in computing robust actions under different contact modes. Consider the shared grasping system in Figure~\ref{fig:example1} as an example. The friction coefficients are estimated to be 1.2 and 0.25 for hand-object and table-object contacts, respectively.

There are 81 contact modes for the four contact points. Algorithm~\ref{alg:shared grasping control} computes 13 modes with positive geometrical stability margin $\Phi_g$. We hand-pick six modes with object motion and solve them with Procedure~\ref{proc:wrench stamping}.
The results are shown in Figure~\ref{fig:exp_all_modes}.

\subsection{Contact Mode Selection and Control}
\label{sub:contact_mode_selection_and_control}
\begin{figure}[ht]
    \centering
    \includegraphics[width=0.45\textwidth]{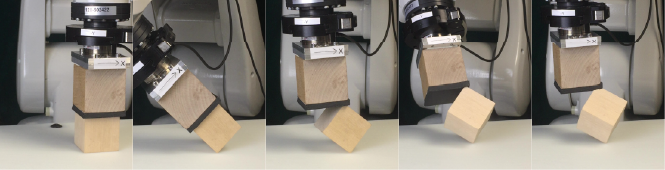}
    \caption{Flipping a cube on its corners.}
    \label{fig:exp_diamond}
\end{figure}
Using the same object and robot, we demonstrate Algorithm~\ref{alg:shared grasping control} in the task of cube rotation with a palm, as shown in Figure~\ref{fig:exp_diamond}. The task is challenging because the object could fall at any time, the control must always be robust. The task has four stages. The goal velocity for the first two stages are:
\begin{equation}
    {\bf G}_1 = [0\ 0\ 1\ 0\ 0\ 0],\ b_{G1} = -0.1.
\end{equation}
\begin{equation}
{\bf G}_2 = \left[\begin{array}{ll}Adj_{WH} & {\bf I}_3\\0\ 0\ 0\ & 0\ 0\ 1 \end{array}\right],
b_{G2} = [0\ 0\ 0\ -0.1]^T.
\end{equation}
First three rows of ${\bf G}_2, b_{G2}$ constrain the object to be static. Stage three and four are the same but in a different plane. Each stage is divided into eight time steps, at which Algorithm~\ref{alg:shared grasping control} decides which mode and what action to take. There is no vision feedback, the object pose is estimated from robot hand pose and has an accumulating error. We execute the four-stage task eight times and have four complete successes. The failures all happen in the last two stages.

\subsection{Finger Placement Optimization} 
\label{sub:finger_placement_optimization}
\begin{figure}[ht]
    \centering
    \includegraphics[width=0.45\textwidth]{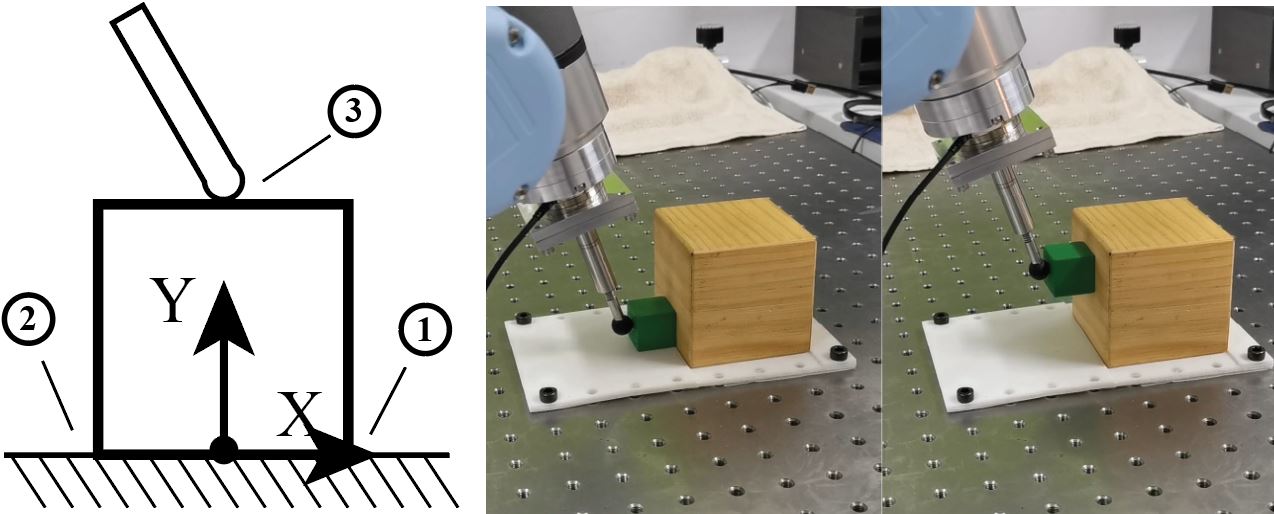}
    \caption{Left: the Finger-block example and the ordering of contacts. Right: Experiment setup. The yellow block is fixed.}
    \label{fig:experiment2}
\end{figure}

\begin{figure}[ht]
    \centering
    \includegraphics[width=0.5\textwidth]{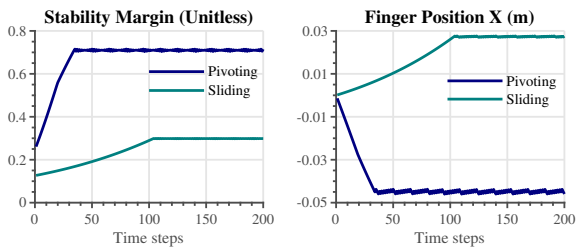}
    \caption{The evolution of finger position and stability margin during the optimization.}
    \label{fig:exp_finger_optimization}
\end{figure}

We test Algorithm~\ref{alg:geometrical parameter optimization} on a point finger manipulation task, as shown in Figure~\ref{fig:experiment2}, left. The width and height of the block are both 0.1m. We optimize finger location for two tasks: 1. pivoting about the left corner with mode ``sff''; 2. Sliding to the left with mode ``llf''. We show the result of 200 iterations of gradient descent in Figure~\ref{fig:exp_finger_optimization}, right, which takes 0.45s in MATLAB. The stability margin increases and reaches a plateau. Note that on the plateau the solution is still zig-zagging by a small magnitude, because the set of contact screws being optimized is varying between iterations, as explained in section~\ref{sub:geometry_optimization}.

\subsection{Compute the Range of Feasible Parameters} 
\label{sub:range_of_feasible_geometries}
\begin{figure}[ht]
    \centering
    \includegraphics[width=0.45\textwidth]{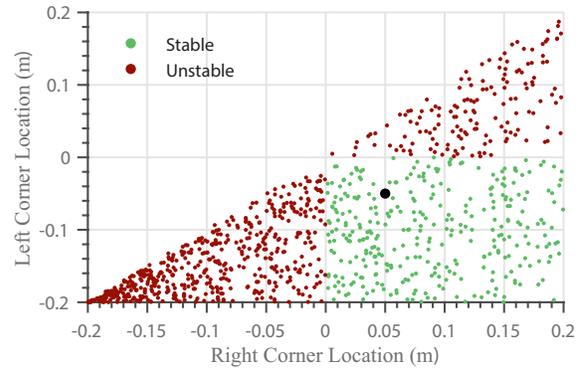}
    \caption{Feasibility of different model parameters. The control is computed for the black dot.}
    \label{fig:experiments_parameter_range}
\end{figure}

Consider again the task of block sliding  (Figure~\ref{fig:experiment2}). Using the optimized finger location and a HFVC computed from Algorithm~\ref{alg:shared grasping control}, we compute the range of table contact locations that produce a positive stability margin for the desired mode ``llf''. We randomly sample 1000 pairs of object right and left corner locations within the range $(-0.2m, 0.2m)$. The result is shown in Figure~\ref{fig:experiments_parameter_range}. A point is marked green if the stability margin is positive.
We can see there is roughly a safety range of $0.05m$ around the nominal point, i.e. the sliding motion can still be successful if the actual environmental contact location changes by 0.05m. We can do the same computation for varying object height, friction coefficient, etc.

\begin{figure}[ht]
    \centering
    \includegraphics[width=0.4\textwidth]{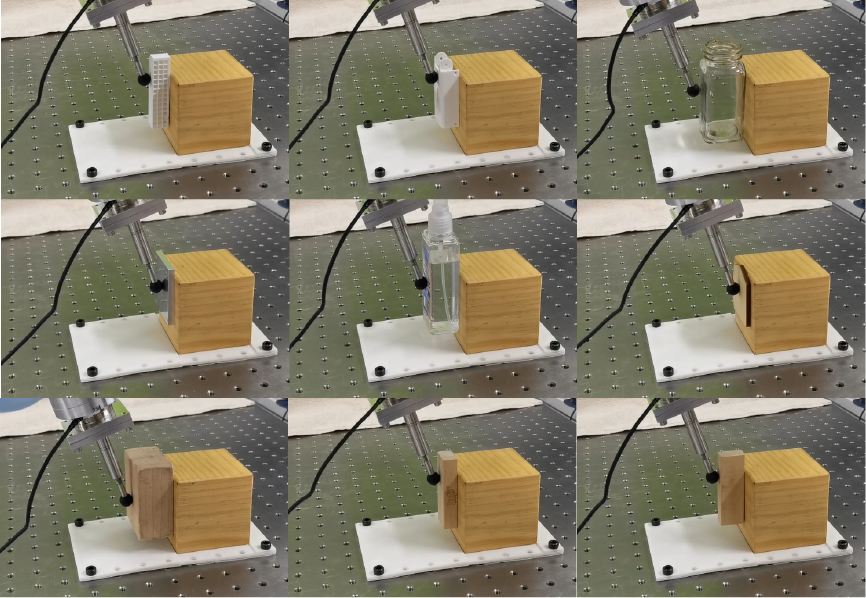}
    \caption{The same control law works for different objects.}
    \label{fig:experiments_stepcrawling_all}
\end{figure}

We implement this sliding action with a wide range of objects and also with gravity as disturbance force. The results are shown in Figure~\ref{fig:experiments_stepcrawling_all}.

\section{CONCLUSION AND DISCUSSION}
\label{sec:conclusion}
To conclude, we provide stability analysis method for shared grasping problems that involves multiple contact modes. We demonstrate how to find the most robust contact mode and compute robot actions to achieve the desired system velocity, as well as how to refine finger placements to improve stability. We are working on generalizing the method to 3D problems. We need to revisit the design of stability margins (\eqref{eq: stability margin} and \eqref{eq:stability margin psi}), which currently relies on 3D geometry.

We demonstrate the robustness of our method in a variety of experiments. The speed of robot motion is limited by the bandwidth of the robot. Recent progress on direct-drive robot hands \cite{Bhatia19Direct} opens possibilities for faster shared grasping.

Although shared grasping is designed for making environmental contacts, the stability analysis and algorithms also work for real in-hand manipulation problems if there are two (groups of) fingers.

Shared grasping complements grasping as another category of manipulations with good robustness. This work shows its potential for contact-rich manipulations that are reliable enough for real-world applications.

\section*{Acknowledgments}
We would like to thank Eric Huang and Xianyi Cheng for insightful discussions. We also want to thank Nikhil Chavan-Dafle and Ankit Bhatia for their suggestions in the writing of the paper.

\bibliographystyle{plainnat}
\bibliography{RSS20PartialGrasp}

\end{document}